\documentclass[10pt,twocolumn,letterpaper]{article}

\usepackage{cvpr}              
\definecolor{cvprblue}{rgb}{0.21,0.49,0.74}
\usepackage[pagebackref,breaklinks,colorlinks,allcolors=cvprblue]{hyperref}

\usepackage[accsupp]{axessibility}  

\usepackage[utf8]{inputenc} 
\usepackage[T1]{fontenc}    
\usepackage{hyperref}       
\usepackage{url}            
\usepackage{booktabs}       
\usepackage{amsfonts}       
\usepackage{nicefrac}       
\usepackage{microtype}      
\usepackage[table,xcdraw]{xcolor}
\usepackage{enumitem}
\usepackage{colortbl}
\usepackage{threeparttable}
\usepackage{multirow}
\usepackage{pifont} 
\usepackage{graphicx}
\usepackage{amsmath} 
\usepackage{stfloats} 
\usepackage{caption}
\usepackage{subcaption}
\usepackage[normalem]{ulem}
\usepackage[most]{tcolorbox} 
\usepackage{makecell}
\usepackage{titletoc} 
\definecolor{keywordcolor}{RGB}{0, 50, 150} 
\definecolor{bracketcolor}{RGB}{100, 100, 100} 

\newcommand{\tablestyle}[2]{\setlength{\tabcolsep}{#1}\renewcommand{\arraystretch}{#2}\centering\footnotesize}
\newcommand{\ptag}[1]{\textcolor{keywordcolor}{$\langle$\textit{#1}$\rangle$}}

\newcommand{\zycnote}[1]{\textcolor{green}{#1}}

\def\paperID{} 

\def\confName{CVPR}
\def\confYear{2026}

\title{\raisebox{-0.15cm}{\includegraphics[height=2em]{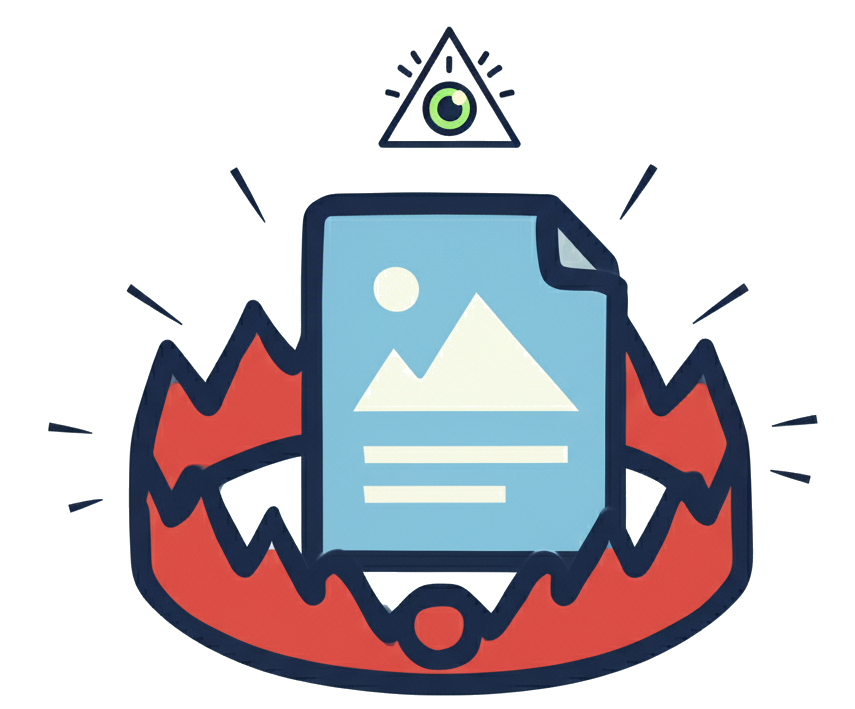}}~The Coherence Trap: When MLLM-Crafted Narratives Exploit Manipulated Visual Contexts}

\author{
\textbf{Yuchen Zhang}$^{\textbf{1}}$ \quad
\textbf{Yaxiong Wang}$^{\textbf{2*}}$ \quad
\textbf{Yujiao Wu}$^{\textbf{3}}$ \quad
\textbf{Lianwei Wu}$^{\textbf{4}}$ \quad
\textbf{Li Zhu}$^{\textbf{1}}$ \quad
\textbf{Zhedong Zheng}$^{\textbf{5}}$\\
$^{1}$School of Software Engineering, Xi'an Jiaotong University \\
$^{2}$School of Computer Science and Information Engineering, Hefei University of Technology \\ 
$^{3}$CSIRO \quad
$^{4}$Northwestern Polytechnical University \quad
$^{5}$University of Macau\\
{\tt\small yczhang@stu.xjtu.edu.cn} \quad
{\tt\small wangyx@hfut.edu.cn}\\
{\small $^{*}$Corresponding author}
}

\begin{document}
\pagestyle{plain}

\twocolumn
[
{
\renewcommand\twocolumn[1][]{#1}
\maketitle
\begin{center}
\setlength{\abovecaptionskip}{8pt} 
\setlength{\belowcaptionskip}{0pt} 
\centering
\vspace{-0.9cm}
\includegraphics[height=2.0in, width=6.8in]{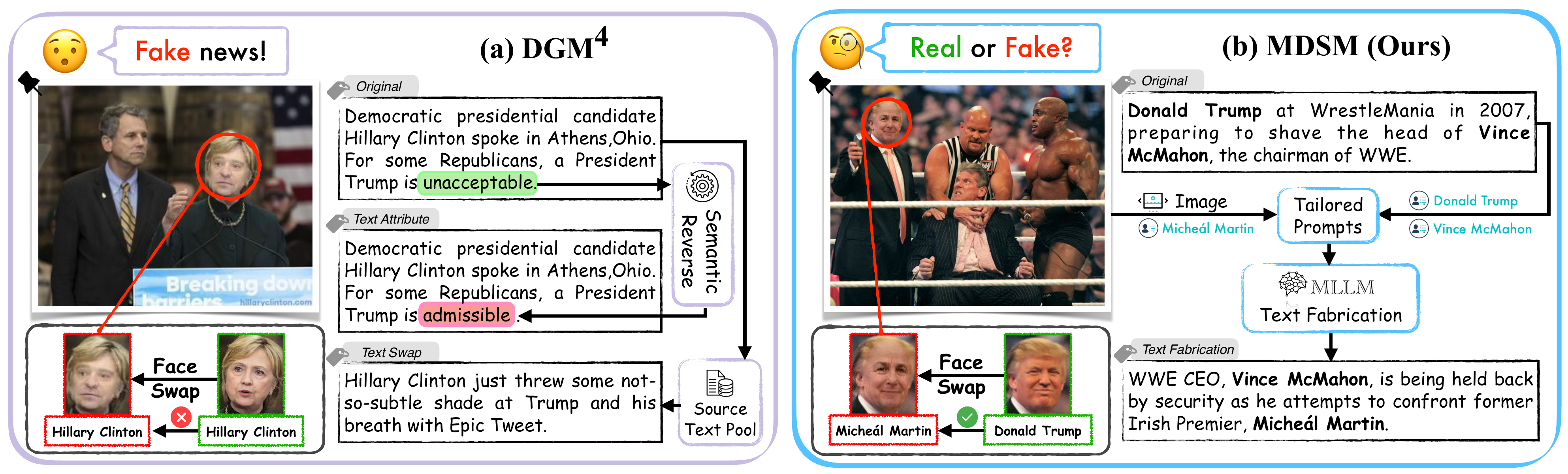}
\vspace{-0.2cm}
\captionof{figure}{{Comparison between widely-used \(\text{DGM}^4\) \citep{DGM4_TPAMI} and MDSM (ours). (\emph{left}) \(\text{DGM}^4\) typically treats visual context manipulation as two independent procedures as rule-based text editing and image editing. The absence of context integration often results in poorly aligned samples, which can be readily perceived by the public. (\emph{right}) In a real-world scenario, we usually face well-wrapped content on purpose. To mimic such a situation, we propose a new manipulation paradigm, which explicitly provides the modified image as well as meta info of facial editing to Multimodal Large Language Model (MLLM). Then we harness MLLM to generate contextually consistent, deceptive texts to form the challenging image-text pairs.}}
\label{fig:motivation}
\end{center}
}
]

\begin{abstract}
The detection and grounding of multimedia manipulation has emerged as a critical challenge in combating AI-generated disinformation. While existing methods have made progress in recent years, we identify two fundamental limitations in current approaches: (1) Underestimation of MLLM-driven deception risk: prevailing techniques primarily address rule-based text manipulations, yet fail to account for sophisticated misinformation synthesized by multimodal large language models (MLLMs) that can dynamically generate semantically coherent, contextually plausible yet deceptive narratives conditioned on manipulated images; (2) Unrealistic misalignment artifacts: currently focused scenarios rely on artificially misaligned content that lacks semantic coherence, rendering them easily detectable. 
To address these gaps holistically, we propose a new adversarial pipeline that leverages MLLMs to generate high-risk disinformation. Our approach begins with constructing the MLLM-Driven Synthetic Multimodal (MDSM) dataset, where images are first altered using state-of-the-art editing techniques and then paired with MLLM-generated deceptive texts that maintain semantic consistency with the visual manipulations. 
Building upon this foundation, we present the \textbf{A}rtifact-aware \textbf{M}anipulation \textbf{D}iagnosis via MLLM (AMD) framework featuring two key innovations: Artifact Pre-perception Encoding strategy and Manipulation-Oriented Reasoning, to tame MLLMs for the MDSM problem. Comprehensive experiments validate our framework's superior generalization capabilities as a unified architecture for detecting MLLM-powered multimodal deceptions. In cross-domain testing on the MDSM dataset, AMD achieves the best average performance, with 88.18 ACC, 60.25 mAP, and 61.02 mIoU scores.  The code is available at \url{https://github.com/YcZhangSing/AMD}.
\end{abstract}
\vspace{-1cm}
    
\section{Introduction}
\label{sec:intro}

\begin{table*}[t]
\caption{Comparison of the proposed MDSM with existing misinformation datasets, where MM Det., Text Det., Man. Type Det., and Im. GD stand for Multi-media Detection, Text Detection, Manipulation Type Detection and Image Grounding.}
\label{tab:datasets_comp}
\centering
\vspace{-0.3cm}
\begin{threeparttable}

\footnotesize
\setlength{\tabcolsep}{1.25mm}
\setlength{\extrarowheight}{-2pt}
{
\resizebox{0.95\textwidth}{!}{\begin{tabular}{@{}lcccccccccc@{}}
\toprule[1.5pt]
\multirow{2}{*}{Datasets} & \multirow{2}*{Samples} &\multicolumn{2}{c}{Modality} &\multicolumn{4}{c}{Tasks} & {Training} &{Semantic} & {MLLM} \\   [-0.05cm] 
\cmidrule(lr){3-4} \cmidrule(lr){5-8}  \\ [-0.25cm] 
& & Text & Image   &MM Det.   &  Text Det. & Man. Type Det. &Im. GD & Support  &Alignment &Inclusion  \\  [-0.05cm] 
\midrule

LIAR ~\citep{Text_news_dataset_liar} &13K & \ding{51} & \ding{55}& \ding{55}& \ding{51}& \ding{55}& \ding{55}& \ding{51} & \ding{55}& \ding{55}\\

DFIM-HQ ~\citep{DFIM_HQ} & 140K & \ding{55} & \ding{51}& \ding{55}& \ding{55}& \ding{51}& \ding{51}& \ding{51} & \ding{55}& \ding{55}\\

MEIR ~\citep{MEIR_ACM18} & 139k & \ding{51} & \ding{51}& \ding{51}& \ding{51}& \ding{55}& \ding{55}&  \ding{51} & \ding{55}& \ding{55}\\

MiRAGeNews ~\citep{huang-etal-2024-miragenews} & 15k & \ding{51} & \ding{51}& \ding{51}& \ding{51}& \ding{55}& \ding{55}&  \ding{51} & \ding{55}& \ding{51}\\


COSMOS ~\citep{COSMOS_AAAI} & 453k & \ding{51} & \ding{51}& \ding{51}& \ding{51}& \ding{55}& \ding{55} & \ding{51} &  \ding{55}& \ding{55}\\

MMFakeBench ~\citep{liu2024mmfakebench} & 11k & \ding{51} & \ding{51}& \ding{51}& \ding{51}& \ding{51}& \ding{55}& \ding{55} & \ding{55}& \ding{55}\\

\(\text{DGM}^4\) ~\citep{DGM4} & 230k & \ding{51} & \ding{51}& \ding{51}& \ding{51}& \ding{51}& \ding{51}& \ding{51} & \ding{55}& \ding{55}\\
\cellcolor{gray!20}MDSM (Ours) & \cellcolor{gray!20}441k & \cellcolor{gray!20}\ding{51} & \cellcolor{gray!20}\ding{51} & \cellcolor{gray!20}\ding{51} & \cellcolor{gray!20}\ding{51} & \cellcolor{gray!20}\ding{51} & \cellcolor{gray!20}\ding{51} & \cellcolor{gray!20}\ding{51} & \cellcolor{gray!20}\ding{51} & \cellcolor{gray!20}\ding{51} \\

\bottomrule[1.5pt]
\end{tabular}}}
    \end{threeparttable}
\vspace{-0.5cm}
\end{table*}

Generative techniques have become a transformative force in artificial intelligence \citep{wu2024deepseekvl2,peng2025lmmr1,dawn2024_ICLR, fu2025speculative,Florence2_CVPR,zhang2025ctrl}, showing remarkable adaptability across various domains and modalities. 
These advancements, while enriching multimedia content, also pose significant challenges to information security. In the media industry in particular, maliciously fake content manipulated by such models can profoundly mislead audiences \citep{Defen_fake_news_NeurIPS,deepfake_eccv_shao,yaxiongAIGCDet4}. 
The unchecked spread of fake media has already negatively affected political, financial, and other sectors \citep{cantarella2023does_fakenews,fake_news_impact_financial,rocha2021impact_health}, gradually becoming a major social issue in current epoch \citep{Olan2024_fakeNewsImpact,FakeSV_yaxiongAIGCDet1,OmniVL_Guard}.

While various misinformation scenarios have been explored including COSMOS \cite{COSMOS_AAAI}, which investigates out-of-context social image-text pairs, and $\text{DGM}^4$ \cite{DGM4}, which focuses on detecting randomly tampered regions or words. MMFakeBench \citep{liu2024mmfakebench} has made preliminary efforts in detecting semantically aligned fake news. It considers fake images synthesized by text-to-image models to maintain semantic consistency with the rumor texts. However, such semantically aligned instances constitute only 30\% of its dataset, and the overall dataset size is merely 11k, which is insufficient to support the training of a robust detection model. 

The above analysis reveals two critical limitations in existing research: 1) Neglect of emerging risks from modern Multi-modal Large Language Models (MLLMs): Current paradigms predominantly address rule-based text manipulation, overlooking the sophisticated linguistic capabilities of modern MLLMs. MLLM-generated text exhibits superior fluency and contextual coherence, significantly increasing deception potential and public susceptibility.  2) Semantic misaligned artifacts.  Most methods independently manipulate visual and textual elements, producing semantically discordant multimedia. This misalignment not only renders manufactured disinformation too simplistic to effectively deceive the public, but also fails to replicate real-world adversarial behavior, as sophisticated attackers typically maintain meticulous visual-textual consistency to maximize manipulative impact. Both limitations render the multi-modal disinformation scenarios considered in existing works insufficiently realistic.


To address these weaknesses, we take the risk of MLLM-driven fabrication into consideration and focus on detecting the semantic-aligned manipulation. We begin by constructing the MLLM-Driven Synthetic Multi-modal (MDSM) dataset, where both images and text are manipulated in a coordinated manner, resulting in semantically aligned image-text pairs. For image manipulation, we consider the typical Face Swap and Face Attribute editing. For text, We innovatively guide MLLM to generate modality-aligned yet misleading fake narratives using image editing metadata. As shown in Fig.~\ref{fig:motivation}(b), after replacing Donald Trump's face with Micheál Martin's, we use the swapped name, Micheál Martin, to guide MLLM in generating text, ensuring that the named entity in the text aligns with the image. Following this strategy, we construct \textbf{over 441k} sample pairs.

The alignment of modalities and the authentic texts from MLLMs pose significant challenges for the detection of manipulated media. First, the strategy of perceiving inconsistencies between images and text through contrastive learning, as employed by prior works \citep{zhang2024asap,DGM4, yaxiongAIGCDet3, yaxiongAIGCDet5}, is ineffective in MDSM where images and text are well-matched already. Merely observing aligned image-text pairs is inadequate for reliable detection. Consequently, external clues and contextual knowledge are essential. Second, existing architectures like ASAP \citep{zhang2024asap} and HAMMER \citep{DGM4}, which feature multiple detection and grounding heads, are complex and lack generalizability to unseen media sources. To address these challenges, we propose \textbf{A}rtifact-aware \textbf{M}anipulation \textbf{D}iagnosis via MLLM (AMD), which leverages MLLMs' comprehensive understanding of real-world multimedia and their ability to provide unified textual outputs. And AMD generates detection and grounding results in a coherent, text-based format, offering a more intuitive and generalized solution. In summary, our contributions of this paper are as follows:

\begin{itemize}[leftmargin=2em,itemsep=0.25em]
  \item We make an early exploration to detect and ground the MLLM-driven manipulation in multimedia and establish an MLLM-Driven Synthetic Multimodal (MDSM) dataset, which defines a more challenging and practical problem for misinformation detecting. 
  \item We propose a strong baseline, \emph{i.e.,} Artifact-aware Manipulation Diagnosis framework (AMD), for the MDSM problem. AMD synergizes artifact pre-perception encoding and manipulation-oriented reasoning to effectively adapt MLLMs for precise manipulation analysis. 
  \item Comprehensive evaluations validate AMD's effectiveness and generalization capability, outperforming existing methods while maintaining parameter efficiency. With only 0.27B parameters, AMD achieves the best domain generalization average performance on both MDSM (88.18 ACC) and $\text{DGM}^4$ (74.47 ACC).
\end{itemize}




\section{MDSM Dataset Construction}

\begin{figure*}[t]
    \centering
    \includegraphics[width=0.95\linewidth]{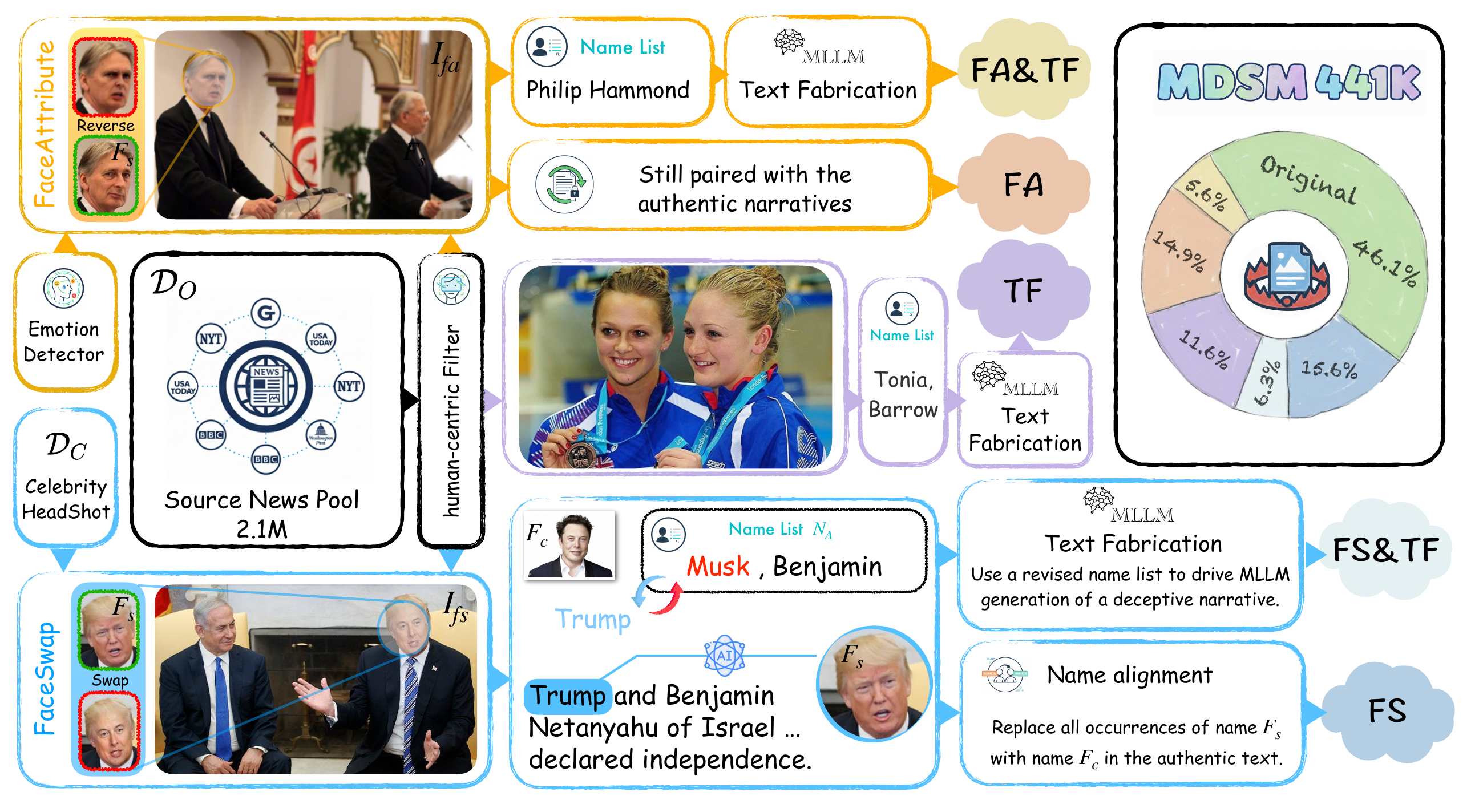}
    \vspace{-0.4cm}

    \caption{
    MDSM Construction Pipeline. \textcolor[rgb]{0.996, 0.678, 0}{$\square$}: The emotion detector helps the face editing model reverse the facial expression. The manipulated image is paired with MLLM-generated fabricated text for the \textit{Face Attribute\&Text Fabrication category (FA\&TF)}, or with authentic text for the \textit{Face Attribute category (FA)} .  
    \textcolor[rgb]{0.757, 0.702, 0.898}{$\square$}: The MLLM fabricated text paired with authentic image for the \textit{Text Fabrication category (TF)}.
    \textcolor[rgb]{0, 0.631, 1}{$\square$}: After swapping the face, the name list is updated for text-image alignment, and the manipulated image is paired with MLLM-generated fabricated text for the \textit{Face Swap\&Text Fabrication category (FS\&TF)}, or with aligned authentic text for the \textit{Face Sap category (FS)}.
    }
    
    \label{fig:dataset}
    \vspace{-0.36cm}
\end{figure*}



As shown in Fig.~\ref{fig:dataset}, the collected source news data undergoes two key synthesis processes: 1) advanced image editing models generate visual manipulations, and 2) MLLMs produce text narratives that are semantically aligned with these visuals. We elaborate on these processes below.


\subsection{Multi-Modal Media Source Collection}

We use the GoodNews ~\citep{GoodNews}, VisualNews ~\citep{VisualNews}, and N24News ~\citep{N24News} datasets as the Source News Pool $ \mathcal{D}_{O} $, 
which consists of over 2.1M image-text pairs sourced from various real-world news outlets. Given the significant influence of human-centric news among various forms of multi-modal media, we focus on human-centric data for MDSM. $ \mathcal{D}_{O} $ is firstly filtered by detecting faces in images with Dlib ~\citep{dlib} and identifying person names in texts with BERT ~\citep{devlin2018bert}. Only pairs, $ p_{s} = (I_{s},T_{s}) $, containing both faces and named entities are used for manipulation. Additionally, we collect the \textit{Celebrity Head-shot Dataset} $\mathcal{D}_{C}$, which contains about 30k pairs of head-shot images and corresponding names to facilitate the aligned manipulation for Face Swap.

\subsection{Multi-Modal Media Manipulation}
In the image modality, two main attacks, Face Swap (FS) and Face Attribute (FA), are employed. For the text modality, we utilize advanced MLLM to generate semantic-aligned Text Fabrication (TF) for the images. Combining these image and text manipulations, we construct a total of five types of multimodal forgeries: FS, FS\&TF, FA, FA\&TF, and TF. The data generation pipeline is illustrated in Fig. \ref{fig:dataset}, and the detailed generation process is described as follows:

\begin{figure*}[t]
    \centering
    \includegraphics[width=0.9\linewidth]{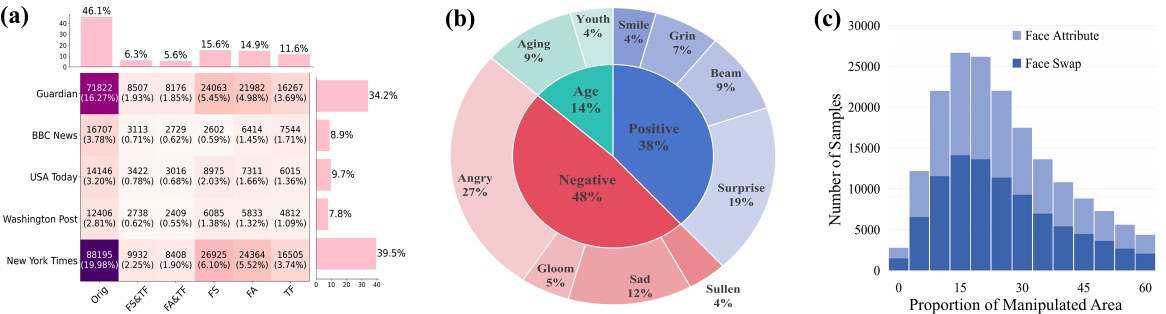}\\
    \vspace{-0.2cm}
    \caption{MDSM Statistics. (a) Distribution of media sources and manipulation categories. (b) Types of face attributes. (c) Distribution of the manipulated image area proportion of the entire image. }
    \label{fig:dataset_stat}
    \vspace{-0.53cm}
\end{figure*}

$\triangleright$  \textbf{Face Swap.} Face swap is a critical tool for attackers to forge images of public figures and politicians, posing threats to societal security. We use two representative face swap methods, SimSwap ~\citep{chen2020simswap} and e4s  ~\citep{liu2022fine_e4s}, to perform such manipulations. We prioritize modifying larger faces to target the primary subject in the image (Fig.~\ref{fig:dataset_stat}c). Given a source image \( I_s \), we randomly choose one of the two methods and replace the largest face \( F_s \) in \( I_s \) with a face \( F_c \) from \( \mathcal{D}_C \), generating a manipulated face swap sample \( I_{fs} \). The bounding box \( y_{box} = \{x_1, y_1, x_2, y_2\} \) of the swapped face and the name of \( F_c \) are recorded. To keep the image-text aligned, corresponding processing is also done on the authentic text. We use MLLM to identify the name of $F_s$, as shown in Fig.~\ref{fig:dataset}, $ F_s$ is identified as Trump. We then refine the authentic text for FS category by replacing this name with $F_c$'s name.

$\triangleright$ \textbf{Face Swap and Text Fabrication.} 
We use Qwen2-VL ~\citep{Qwen2-VL} to generate consistent but misleading narratives. This requires knowing: 1) the inserted person's name, and 2) the names of people remaining in \( I_{fs} \). Using the same strategy as above, we get $F_s$' name. We extract the full name list from the original text \( T_s \) using BERT ~\citep{devlin2018bert}, and then replace $F_s$ with inserted person's name to form the final name list \( N_A \). Finally, we input \( I_{fs} \) and \( N_A \) into Qwen2-VL to generate aligned text.

$\triangleright$ \textbf{Face Attribute.} 
Face emotion editing is also considered in our dataset. Our pipeline uses StyleCLIP  ~\citep{patashnik2021styleclip} and HFGI  ~\citep{wang2022high_HFGI} for attribute manipulations. Firstly, we analyze facial expressions using an emotion detector ~\citep{Arriaga2017RealtimeCN} to determine positive or negative emotions. We then randomly select a method to manipulate the primary face \( F_s \)' attributes inversely to the classification outcome, producing \( I_{fa} \). To ensure diversity, we control manipulation intensity with variable prompts and introduce age modifications. The distribution of face attribute prompts is shown in Fig.~\ref{fig:dataset_stat}(b), with \( y_{box} \) stored as annotation. Since the characters in \( I_{fa} \) have not changed, the paired text in this category is still authentic.

$\triangleright$ \textbf{Face Attribute and Text Fabrication.}
Similar to face swapping, text forgery for face attribute editing is also generated by Qwen2-VL but with distinct prompts. Specifically, we instruct the MLLM to focus primarily on facial expressions to generate narratives that conform to the characters' demeanor. The input full name list is initially extracted from the source text $T_s$.

$\triangleright$ \textbf{Text Fabrication.} For the TF category, we also use BERT ~\citep{devlin2018bert} to extract the name list \( N_A \) from the original text. Then, we input \( N_A \) and the original image into the MLLM to generate narratives that match the implied meaning but are still fabricated.

\subsection{Dataset Statistics}

With the above steps, we finally harvest our MDSM dataset $\mathcal{D}_M$, a large-scale, high semantic-aligned multi-modal benchmark with high-fidelity texts from MLLM. The distribution of manipulation categories is well balanced and consistent with previous datasets, ensuring fair evaluation across manipulation modes (Fig.~\ref{fig:dataset_stat}a). Compared with the existing manipulation detection benchmarks in Tab.~\ref{tab:datasets_comp},  MDSM has the following advantages: \textbf{{1)Risk Consideration of MLLM.}} MDSM acknowledges the emerging challenges posed by MLLMs and utilizes multi-modal methods to create semantically coherent and contextually plausible narratives for manipulated images. This scenario, though underexplored, is a highly significant and timely problem in the modern large model era. \textbf{2)Semantic Alignment.} MDSM is a high aligned multi-modal media manipulation benchmark, which is a significant and more practical scenario for multi-modal manipulation detection. \textbf{3)Large Scale.} Our MDSM comprises 441,423 samples and is the largest benchmark for detecting and grounding multi-modal manipulation. \textbf{4)Diverse Multi-media Sources.} The multi-modal media of MDSM sources from diverse media sites, including \textit{The Guardian}, \textit{The New York Times}, \textit{The Washington Post}, \textit{USA Today}, and \textit{the BBC}. Consequently, the generality of methods can be assessed via cross-domain evaluation.

Our proposed MDSM defines three tasks: \textbf{1)Fake Multi-modal Media Detection}. True for the manipulated media and False for the original ones. \textbf{2)Manipulation Type Detection}, recognizing  Face Swap (FS), Face Attribute (FA), and Text Fabrication (TF). \textbf{3)Image Grounding}, locating the bounding box of the manipulated region in image.

\section{Methodology}
\label{sec:method}
\begin{figure*}[t]
    \centering
\includegraphics[width=0.97\linewidth]{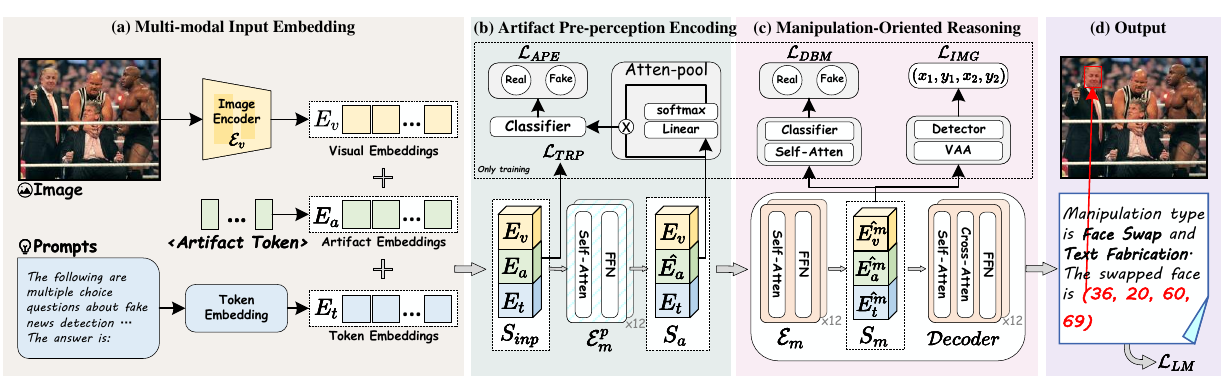}\\
    \vspace{-0.35cm}
    \caption{Overview of the proposed AMD framework. (a) Maps the manipulated image and prompts into a unified representation, incorporating an artifact token. (b) Utilizes the artifact pre-perception multimodal encoder $\mathcal{E}_m^p$ to extract perceptual clues. (c) Processes multi-modal features through $\mathcal{E}_m$ to generate text-based detection results. (d) Outputs and visualizes the final manipulation analysis.}
    \label{fig:method}
    \vspace{-0.45cm}
\end{figure*}

Fig.~\ref{fig:method} illustrates the architecture of our Artifact-aware Manipulation Diagnosis framework (AMD). Built upon Florence-2 ~\citep{Florence2_CVPR} to leverage real-world knowledge, AMD follows a sequence-to-sequence framework for joint textual detection and grounding. Multimodal inputs are processed through three stages, and outputs localized predictions with textual explanations.

\subsection{Multi-modal Input Embedding.}
\noindent \textbf{Prompt construction.} To adapt the MLLM for the MDSM task while preserving its inherent knowledge, we develop heuristic question(human)-answer(assistant) prompts:
\begin{tcolorbox}[
    colback=gray!5,
    colframe=gray!40,
    title=\textbf{Prompt Template Definition},
    arc=2mm,
    fonttitle=\bfseries\small,
    boxrule=0.5pt,
    left=2mm, right=2mm, 
    top=1mm, bottom=1mm, 
    middle=1mm,          
    toptitle=1mm, bottomtitle=1mm, 
    fontupper=\footnotesize 
]
\noindent\texttt{\textbf{\#\#\#Human:}}\ptag{Task}\ptag{Options}\ptag{Grounding}
    \\[0.1em] 
    \noindent\texttt{\textbf{\#\#\#Assistant:}}\ptag{Response}\textcolor{bracketcolor}{[}\ptag{Coordinates}\textcolor{bracketcolor}{]}
        \tcbline 
    \begin{itemize}[nosep, leftmargin=*, labelsep=0.5em] 
        \item \ptag{Task}: Specifies manipulation detection objective \& pairs input.
        \item \ptag{Options}: Lists all candidate answers for the MDSM task.
        \item \ptag{Grounding}: Triggers region localization conditionally.
        \item \ptag{Response}: Encapsulates the correct answers.
        \item \textcolor{bracketcolor}{[}\ptag{Coordinates}\textcolor{bracketcolor}{]}: Encloses tamperd region coordinates.
    \end{itemize}
\end{tcolorbox}


\noindent \textbf{Artifact Token Embeddings.} To effectively adapt the MLLM into MDSM context while preserving its pretrained knowledge, we introduce a learnable Artifact Token that explicitly encodes artifacts from heterogeneous inputs. Formally, let the artifact token embeddings be denoted as  ${E}_a \in \mathbb{R}^{n_a \times d}$, where  $n_a$  indicates the token count and $d$ the embedding dimension. The textual input is processed through an embedding layer to obtain text embeddings  ${E}_t \in \mathbb{R}^{n_t \times d}$ , while the visual input is encoded via a vision backbone  $\mathcal{E}_v$  followed by a LayerNorm-augmented linear projection, yielding image embeddings $ {E}_v \in \mathbb{R}^{n_v \times d}$. The above embeddings are concatenated to construct the input sequence: ${S}_{\text{inp}}=[{E}_v; {E}_a; {E}_t]$, where  $[\cdot; \cdot]$  means  concatenating along the token dimension.

\subsection{Artifact Pre-perception Encoding}
This stage aims to perceive manipulation artifacts within input data and condense these forensic clues into the artifact token. Specifically, the input sequence $S$ undergoes processing through the artifact pre-perception multimodal encoder $\mathcal{E}^p_m$, yielding \( \hat{S}= [\hat{E}_v; \hat{E}_a; \hat{E}_t] \). To inject artifact-aware clues into the artifact token embedding $\hat{E}_a$, we pick $\hat{E}_a$ from $\hat{S}$ and feed it into an artifact-aware classification head. As illustrated in Fig.~\ref{fig:method}, this classification head is optimized via a manipulation detection objective to explicitly encode artifact-related patterns into $\hat{E}_a$.Then 


Particularly, the embedding $\hat{E}_a$ is encoded into a global representation via weighted pooling. Firstly, the token scores $ \mathcal{W} \in \mathbb{R}^{1 \times n_a} $ are calculated as:
\begin{equation}
\label{attn1}
\mathcal{W} = m^{\top} \text{ReLU}(\mathcal{M}\hat{E}_a^{\top} +b ),
\end{equation}
where \( \mathcal{M} \in \mathbb{R}^{h \times d} \), \( m \in \mathbb{R}^{h} \), and \( b \in \mathbb{R}^{h} \), with \( h \) as the hidden dimension. After normalizing \( \mathcal{W} \) via softmax, the artifact representation \( \bar{E}_a  \) is derived as a weighted sum, $\text{softmax}(\mathcal{W}) \cdot  \hat{E}_a$.

Then we equip a binary classifier \(C_{a}\) to determine whether artifact traces are present:
\begin{equation}
\label{apeloss}
    \mathcal{L}_{APE} = \mathbb{E}_{(I,T) \sim \mathcal{D}_M} \mathbf{CE}(C_{a}(\bar{E}_a),y_{fd}),
\end{equation}
where $\mathbf{CE}$ means cross-entropy loss, $y_\text{fd}$ is the label of fake multimodal media detection task.

\noindent\textbf{Task Adaption \& Knowledge Preservation.}  To effectively inject the artifact clues into $\hat{E}_a$ without distorting the original real-world knowledge of MLLM,  two strategies are adopted, we 1) freeze the parameters of $\mathcal{E}_m^p $ during artifact perception loss optimization (Eq.~\ref{apeloss}), such that allowing more artifact clues can be accumulated into the artifact token as well as preserving the raw MLLM knowledge; 2) replace the text and image embeddings in $\hat{S}$ with the original ones to preserve the original MLLM knowledge, \emph{i.e.,} feeding $ S_a= [{E}_v; \hat{E}_a; {E}_t]$ to the subsequent modules, as shown in Fig.~\ref{fig:method}.

\subsection{{Manipulation-Oriented Reasoning}}
Manipulation-Oriented Reasoning (MOR) targets to generate the textual answer in response to the question prompt. 
To acquire an accurate response, we augment the network optimization in MOR with two guiding tasks: visual Artifact Capture via Grounding and Manipulation-focused Guidance.


\noindent\textbf{Visual Artifact Capture via Grounding.} 
The sequence $S_a$ is fed into multimodal encoder $\mathcal{E}_m$, resulting in a new sequence  \( S_m = [\hat{E}_v^m; \hat{E}_a^m; \hat{E}_t^m ]\).  Given that visual embeddings contain rich local spatial information related to artifact traces, we propose a \textit{Visual Artifact Aggregation} (VAA) module to aggregate spatial information in \(\hat{E_v^m}\) to perform manipulation bbox grounding. Firstly, the \(\hat{E_a^m}\) is transformed into a query token \(q_a \in \mathbb{R}^{1\times d}\) using the attention-based weighted pooling (Eq.~\ref{attn1}). Then, $q_a$ collects visual manipulation clues from image features \(\hat{E_v^m}\) via cross attention:
\begin{equation}
    u_{agg} = \text{Attention}(q_a, \hat{E}_{v}^m, \hat{E}_{v}^m).
\end{equation}
Subsequently, the \(u_{agg}\) is sent to the bbox detector to generate artifact coordinates. We follow~\cite{Loss_IoU} to construct the image manipulation grounding loss using  L1 loss \(\mathcal{L}_1\) and GIoU loss \(\mathcal{L}_{IoU}\):
\begin{equation}
    \mathcal{L}_{IMG} = \mathbb{E}_{(I,T) \sim \mathcal{D}_M}(\mathcal{L}_1+\mathcal{L}_{IoU}).
\end{equation}

\noindent\textbf{Manipulation-focused Guidance} 
further highlight whether the multimodal input is manipulated or not, tuning the MLLM to be sufficiently sensitive to the fake multi-modal media. To fully capture manipulation-related information embedded within different modalities, we propose a Dual-Branch Manipulation guidance strategy. Specifically, each modality feature in the encoder output sequence \( S_m \) is treated as a query \( Q \) and undergoes interaction for binary classification. Given that artifact traces predominantly appear in the image modality, the sequence composed of \( \hat{E}_a^m \) and \( \hat{E}_v^m \) is regarded as the image modality feature. The interaction process is formulated as:

\begin{equation}
\begin{aligned}
    u_v &= \text{Attention}(\hat{E}_{v+a}^m, \hat{E}_{t}^m, \hat{E}_{t}^m), \\
    u_t &= \text{Attention}(\hat{E}_{t}^m, \hat{E}_{v+a}^m, \hat{E}_{v+a}^m),
\end{aligned}
\end{equation}
where \( \hat{E}_{v+a}^m \) represents the concatenation of \( \hat{E}_{v}^m \) and \( \hat{E}_{a}^m \), while \( \hat{E}_{t}^m \) corresponds to the textual sequence. The cross-modal interaction outputs, \( u_v \) and \( u_t \), are {respectively processed} by a binary classifier \(C_{m}\) to distinguish between manipulated and original multimodal media. Thus the Dual-Branch Manipulation guidance loss can be calculated as:
\begin{equation}
    \mathcal{L}_{DBM} = \mathbb{E}_{(I,T) \sim \mathcal{D}_M}\sum_{x \in \{v, t\}}\mathbf{CE}(C_{m}(u_x), y_{fd}).
\end{equation}

\noindent\textbf{Language modeling.}
The input sequence \(S_a\) is processed through an encoder-decoder architecture, ultimately generating a pure text output that includes choices and coordinates (Fig.~\ref{fig:method}d) as specified in the prompts. In this stage, an autoregressive approach is adopted, where the decoder generates the target sequence \(y\) conditioned on $S_m$. The language modeling loss \(\mathcal{L}_{LM} \) \citep{Florence2_CVPR} is used to supervise the text outputs.

\subsection{Token Redundancy Penalty}
To suppress the redundancy and increase the information density among tokens in \(E_a\), we design a Token Redundancy Penalty (TRP) optimization term.
Specifically, we first encourage the columns of $E_a$ to be as orthogonal as possible by introducing a loss term $\mathcal{L}_{\mathrm{orth}}$, which increases the matrix rank. We construct Gram matrix of $E_a$, $G = E_a E_a^{\top} \in \mathbb{R}^{n_a \times n_a}$, and the orthogonality of the columns can be measured by the off-diagonal elements of the Gram matrix. Ideally, if the columns are orthogonal, the off-diagonal entries of $G$ should be zero. To this end, we define following constraint:
\begin{equation}
\mathcal{L}_{\mathrm{orth}} = \left\| G - \mathrm{Diag}\left(\mathrm{diag}(G)\right) \right\|_F^2,
\end{equation}
where $\mathrm{Diag}(G)$ denotes a diagonal matrix retaining only the diagonal elements of $G$, and $\left\| \cdot \right\|_F$ denotes the Frobenius norm used to aggregate the differentiable loss.

To avoid a potential checkerboard pattern in $E_a$ under the constraint of $\mathcal{L}_{\mathrm{orth}}$—which could lead to loss of information—we further introduce a modulation constraint $\mathcal{L}_{\mathrm{mod}}$ based on the Kullback–Leibler (KL) divergence. Particularly, we first normalize the components to form a distribution: $p_{t,i} = \frac{{E_a}_{t,i}^2}{\sum_{i=1}^{d} {E_a}_{t,i}^2}$. While the target distribution is set as the even distribution (\(\frac{1}{d}\)), thereby encouraging each component to contain information evenly with following constrain: 
\begin{equation}
\mathcal{L}_{\mathrm{mod}} = \frac{1}{n_a} \sum_{t=1}^{n_a} \left( \sum_{i=1}^{d} p_{t,i} \log p_{t,i} + \log d \right).
\end{equation}

Finally, the overall Token Redundancy Penalty is defined as the combination of both terms:
\begin{equation}
\mathcal{L}_{{TRP}} = \mathcal{L}_{\mathrm{orth}} + \mathcal{L}_{\mathrm{mod}}.
\end{equation}

\noindent\textbf{Training.} 
All guiding losses above and the language modeling loss are incorporated into the training process, forming a unified optimization framework as follows:
\begin{equation}
\mathcal{L} = \mathcal{L}_{APE} + \mathcal{L}_{DBM} + \mathcal{L}_{IMG} + \mathcal{L}_{TRP} + \mathcal{L}_{LM}.
\end{equation}

\noindent\textbf{Inference.} All auxiliary heads for  $\mathcal{L}_{APE}, \mathcal{L}_{DBM}, \mathcal{L}_{IMG}$, and $\mathcal{L}_{TRP}$ are discarded during inference. 
For a piece of multimodal media, the image and the question (text \& prompts) follow the same steps shown in Fig.~\ref{fig:method} and generate the textual detection and grounding results.

\section{Experiment}
\label{sec:Experiments}

\begin{table*}[t]
\caption{Comparison of multi-modal learning methods on MDSM, where the background \colorbox{lightgray!25}{gray} indicates the intra-domain performance. The better results in each group are in \textbf{bold}. AVG refers to the average performance across five news domains.}
\label{tab:baselines_Comparison}
\vspace{-0.3cm}
\centering
\begin{threeparttable}
\footnotesize
\setlength{\tabcolsep}{1mm}
\renewcommand{\arraystretch}{1.3} 
{\resizebox{\textwidth}{!}{\begin{tabular}{@{}l|ccccccccccccccccccc@{}}
\toprule[1.5pt]
\multirow{3}{*}{\rotatebox{90}{\textbf{Setting}}} & \multirow{3}*{\textbf{Method}} &\multicolumn{18}{c}{\textbf{Test Domain}} \\ 

\cline{3-20}
 &  & \multicolumn{3}{c}{NYT} & \multicolumn{3}{c}{Guardian} & \multicolumn{3}{c}{USA} & \multicolumn{3}{c}{Wash.} & \multicolumn{3}{c}{BBC} & \multicolumn{3}{c}{\textbf{AVG}}\\ [-0.05cm] 
\cmidrule(lr){3-5} \cmidrule(lr){6-8} \cmidrule(lr){9-11} \cmidrule(lr){12-14} \cmidrule(lr){15-17} \cmidrule(lr){18-20} \\[-0.35cm] 
 & & ACC & mAP & mIoU & ACC & mAP & mIoU & ACC & mAP & mIoU& ACC & mAP & mIoU & ACC & mAP & mIoU & ACC & mAP & mIoU \\[-0.05cm] 

 \midrule
 \multirow{7}{*}{\rotatebox{90}{Zero-Shot}} 

& LLaVA-v1.6-13B ~\citep{LLAVA_NeurIPS}   
& 32.10 & 21.86 & 0.00 & 21.76 & 19.18 & 0.00 & 17.80 & 12.76 & 0.00 & 15.76 & 25.47 & 0.00 & 22.70 & 17.38 & 0.00 & 22.02 & 19.33 & 0.00 \\

& DeepSeek-VL2-27B ~\citep{wu2024deepseekvl2}   
& 38.12  & 21.79  & 0.45  & 30.35  & 20.12  & 0.76  & 22.49  & 37.27  & 0.43  & 21.73  & 20.02  & 1.78  & 29.70  & 26.58  & 1.12 & 28.48  & 25.16  & 0.91\\

 & Yi-VL-34B ~\citep{yiAL2024}    & 45.74  & 24.19  & 0.07  & 34.98  & 23.98  & 0.13  & 20.63  & 36.74  & 0.00  & 19.28  & 19.10  & 0.00  & 31.39  & 26.97  & 0.00 & 30.40  & 26.20  & 0.04\\

 & Qwen2.5-VL-72B ~\citep{bai2025qwen25}   & 47.74  & 29.24  & 0.00  & 35.18  &25.70  & 0.00  & 24.66  & \textbf{40.60}  & 0.00  & 25.11  & 40.29  & 0.28  & 35.89  & 31.51  & 0.00 & 33.72  & 33.47  & 0.06\\

 & Qwen3-VL-235B ~\citep{Qwen3_blog}  & 45.29  & 25.01  & 0.71  & 38.12  & 27.41  & 1.19  & 22.87  & 39.17  & 0.10  & 22.87  & \textbf{40.77}  & 1.34  & 37.22  & \textbf{31.60}  & 0.98  & 33.27  & \textbf{33.69}  & 0.86  \\
 
 & GPT-4o ~\citep{hurst2024gpt4o}& {48.48}  & 27.90  & 0.82  & 35.68  & \textbf{29.49}  & \textbf{1.23}  & 24.62  & 39.88  & 1.37  & 23.62  & 38.89  & 1.22  & 37.19  & 30.48  & 1.20  & 33.92  & 33.33  & 1.17  \\
 
 & Gemini-2.0 ~\citep{team2023gemini} &  \textbf{56.05}  & \textbf{33.16}  & \textbf{1.44}  & \textbf{41.26}  & 24.37  & 1.12  & \textbf{29.60}  & 38.29  & \textbf{1.40}  & \textbf{29.15}  & 35.20  & \textbf{2.42}  & \textbf{38.12}  & 29.13  & \textbf{2.25}  & \textbf{38.83}  & 32.03  & \textbf{1.72}  \\

 \midrule

\multirow{5}{*}{\rotatebox{90}{Tr. on NYT}} & ViLT ~\citep{ViLT_ICML}  & \cellcolor{lightgray!25}83.27  & \cellcolor{lightgray!25}64.27  & \cellcolor{lightgray!25}22.73  & 72.18  & 31.76  & 20.21  & 70.34  & 36.45  & 21.48  & 65.71  & 36.23  & 17.56  & 74.33  & 36.10  & 19.36  & 73.17  & 40.96  & 20.27 \\

 & HAMMER ~\citep{DGM4}   & \cellcolor{lightgray!25}79.20  & \cellcolor{lightgray!25}55.86  & \cellcolor{lightgray!25}51.34  & 68.23  & 40.10  & 21.56  & 71.52  & 41.17  & 13.74  & 68.50  & 41.47  & 13.92  & 67.37  & 42.23  & 16.12  & 70.96  & 44.16  & 23.34 \\
 
 & HAMMER++ ~\citep{DGM4_TPAMI}   & \cellcolor{lightgray!25}79.61  & \cellcolor{lightgray!25}57.06  & \cellcolor{lightgray!25}54.44  & 66.99  & 38.07  & 17.34  & 67.18  & 37.58  & 10.76  & 66.28  & 37.97  & 10.88  & 66.12  & 37.82  & 13.68  & 69.23  & 41.70  & 21.42\\
 
 & FKA-Owl ~\citep{liu2024fkaowl}  & \cellcolor{lightgray!25}\textbf{94.67}  & \cellcolor{lightgray!25}78.18  & \cellcolor{lightgray!25}55.81  & 77.20  & 46.88  & 43.67  & 78.00  & 44.45  & 50.73  & 75.49  & 50.83  & 43.53  & 84.65  & \textbf{60.73}  & 43.28   & 81.60  & 56.77  & 46.23 \\
 
 & \textbf{AMD (Ours)}   & \cellcolor{lightgray!25}92.24 &\cellcolor{lightgray!25}\textbf{84.47}  &\cellcolor{lightgray!25}\textbf{72.94}  &\textbf{80.21}  &\textbf{64.00}  &\textbf{62.51}  &\textbf{78.56}  &\textbf{68.49}  &\textbf{55.17}  &\textbf{82.64}  &\textbf{69.41}  &\textbf{56.66}  &\textbf{86.14}    &60.58  &\textbf{70.54} & \textbf{83.96}  &\textbf{69.39}  &\textbf{63.56} \\
 
\midrule

\multirow{5}{*}{\rotatebox{90}{Tr. on Guardian}} & ViLT ~\citep{ViLT_ICML}   & 68.80  & 43.99  & 21.77  &\cellcolor{lightgray!25} 85.29  &\cellcolor{lightgray!25} 67.34  & \cellcolor{lightgray!25}41.80  & 70.34  & 46.24  & 37.68  & 78.61  & 47.17  & 38.13  & 80.00  & 44.79  & 38.97 & 76.61  & 49.90  & 35.67\\

 & HAMMER ~\citep{DGM4}   & 61.89  & 37.98  & 18.84  & \cellcolor{lightgray!25}78.50  & \cellcolor{lightgray!25}52.40  &\cellcolor{lightgray!25} 51.53  & 74.78  & 50.76  & 43.40  & 75.11  & 50.34  & 46.36  & 81.32  & 50.15  & 56.03  & 74.32  & 48.33  & 43.23 \\
 
 & HAMMER++ ~\citep{DGM4_TPAMI} & 62.75  & 36.45  & 23.76  & \cellcolor{lightgray!25}80.95  & \cellcolor{lightgray!25}59.92  & \cellcolor{lightgray!25}64.67  & 75.36  & 48.77  & 47.13  & 76.30  & 49.56  & 48.91  & 80.12  & 50.36  & 57.97 & 75.10  & 49.01  & 48.49 \\
 
 & FKA-Owl ~\citep{liu2024fkaowl}   & 80.60  & 40.44  & 26.33  & \cellcolor{lightgray!25}\textbf{92.60}  & \cellcolor{lightgray!25}78.24  & \cellcolor{lightgray!25}71.04  & 80.90  & 51.80  & 50.93  & 78.88  & 51.62  & 50.88  & 87.61  & \textbf{68.57}  & 61.82 &  84.12  & 58.13  & 52.20 \\
 
 & \textbf{AMD (Ours)}   & \textbf{84.29}  & \textbf{48.54}  & \textbf{52.38}  & \cellcolor{lightgray!25}91.43  &\cellcolor{lightgray!25} \textbf{80.85}  & \cellcolor{lightgray!25}\textbf{85.09}  & \textbf{88.80}  & \textbf{53.05}  & \textbf{52.51}  & \textbf{86.64}  & \textbf{54.07}  & \textbf{53.27}  & \textbf{89.74}  & 64.75  & \textbf{61.82}  & \textbf{88.18}  & \textbf{60.25}  & \textbf{61.02} \\
 
\bottomrule[1.5pt]
\end{tabular}}}
\vspace{-0.1cm}
\end{threeparttable}

\end{table*}

\begin{table*}[t]
\caption{Comparison of multi-modal learning methods on $\text{DGM}^4$, where the guardian domain with background \colorbox{lightgray!25}{gray} is intra-domain. $\text{P}_{tok}$ is Precision of fake token grounding.}
\label{tab:DGM4_baselines_Comparison}
\vspace{-0.3cm}
\centering
\begin{threeparttable}
\footnotesize
\setlength{\tabcolsep}{1mm}
\renewcommand{\arraystretch}{1.3} 
{\resizebox{\textwidth}{!}{\begin{tabular}{@{}ccccccccccccccccccccc@{}}
\toprule[1.5pt]
 \multirow{3}*{\textbf{Method}} &\multicolumn{20}{c}{\textbf{Test Domain}} \\ 

\cline{2-21}

 & \multicolumn{4}{c}{Guardian} & \multicolumn{4}{c}{USA} & \multicolumn{4}{c}{Wash.} & \multicolumn{4}{c}{BBC} & \multicolumn{4}{c}{\textbf{AVG}}\\ [-0.05cm] 
\cmidrule(lr){2-5} \cmidrule(lr){6-9} \cmidrule(lr){10-13} \cmidrule(lr){14-17} \cmidrule(lr){18-21}  \\[-0.35cm] 
 & ACC & mAP  & $\text{P}_{tok}$& mIoU & ACC  & mAP & $\text{P}_{tok}$& mIoU& ACC & mAP & $\text{P}_{tok}$ & mIoU& ACC & mAP  & $\text{P}_{tok}$& mIoU& ACC & mAP & $\text{P}_{tok}$& mIoU  \\[-0.05cm] 

 \midrule

ViLT ~\citep{ViLT_ICML}&\cellcolor{lightgray!25}68.27 &\cellcolor{lightgray!25}42.29 &\cellcolor{lightgray!25}69.87 &\cellcolor{lightgray!25}43.19 & 52.79&31.28&62.11&33.78&55.76&33.26&57.17&31.10&44.14&39.68&59.06&21.96&55.24&36.63&62.05&32.49 \\

HAMMER ~\citep{DGM4}&\cellcolor{lightgray!25}78.34  &\cellcolor{lightgray!25} 66.79  & \cellcolor{lightgray!25}78.27  &\cellcolor{lightgray!25} 61.09  & 64.97  & 40.49  & 73.76  & 40.51  & 63.54  & 40.26  & 76.13  & 38.53  & 54.97  & 40.84  & 81.48  & 43.74  & 65.45  & 47.10  & 77.41  & 45.97   \\ 
        
HAMMER++ ~\citep{DGM4_TPAMI}&\cellcolor{lightgray!25}  79.13  &\cellcolor{lightgray!25} 67.11  &\cellcolor{lightgray!25} 78.24  &\cellcolor{lightgray!25} 62.15  & 65.25  & 40.74  & 73.24  & 41.14  & 63.83  & 40.34  & 76.17  & 38.21  & 54.24  & 41.25  & 81.73  & 43.23  & 65.61  & 47.36  & 77.34  & 46.19   \\ 
        
FKA-Owl ~\citep{liu2024fkaowl}&\cellcolor{lightgray!25}82.97  &\cellcolor{lightgray!25} 53.86  &\cellcolor{lightgray!25} \textbf{87.70}  &\cellcolor{lightgray!25} 65.69  & 67.57  & 38.97  & \textbf{79.44}  & 32.57  & 67.05  & 37.70  & \textbf{81.55}  & 31.86  & 70.26  & 40.20  & \textbf{84.54}  & \textbf{46.48}  & 71.96  & 42.68  & \textbf{83.31}  & 44.15   \\ 
        
 \textbf{AMD (Ours)}&\cellcolor{lightgray!25}\textbf{84.61}  &\cellcolor{lightgray!25} \textbf{68.50}  &\cellcolor{lightgray!25} 82.78  &\cellcolor{lightgray!25} \textbf{81.24}  & \textbf{70.62}  & \textbf{43.20}  & 75.73  & \textbf{41.99}  &\textbf{ 70.28}  & \textbf{43.36}  & 77.76  & \textbf{39.05}  & \textbf{72.37}  & \textbf{56.57}  & 83.76  & 45.20  & \textbf{74.47}  & \textbf{52.91}  & 80.01  & \textbf{51.87}  \\ 

\bottomrule[1.5pt]
\end{tabular}}}
\end{threeparttable}
\vspace{-0.5cm}
\end{table*}


\begin{table*}[t]
\caption{
Ablation on (a) each component and (b) discussion regarding performance on test set of different MLLMs \& manipulation types.
}
\vspace{-0.03cm}

\label{tab:ablation_and_discussion}
\centering

\begin{subtable}[t]{0.5\textwidth}
  \centering
  \label{tab:cmal}
  \vspace{0.7cm}
  \tablestyle{2pt}{1.0}\scriptsize
  \begin{tabular}{@{}ccccccccccc@{}}
    \multicolumn{5}{c}{Components}
      & \multicolumn{3}{c}{NYT}
      & \multicolumn{3}{c}{Guardian} \\[-0.08cm]
    \cmidrule(lr){1-5} \cmidrule(lr){6-8} \cmidrule(lr){9-11} \\[-0.35cm]
    LM & APE & IMG & DBM & TRP & ACC & mAP & mIoU & ACC & mAP & mIoU \\[-0.08cm]
    \midrule[1.2pt]
    \ding{51} &           &           &           &          &76.92 & 46.38 & 58.77 & 82.33 & 53.57 & 53.55 \\
    \ding{51} & \ding{51} &           &           &          &82.93 & 47.12 & 60.13 & 87.68 & 56.51 & 57.91 \\
    \ding{51} & \ding{51} & \ding{51} &           &          &82.97 & 47.18 & 61.78 & 87.99 & 56.43 & 60.53 \\
    \ding{51} & \ding{51} & \ding{51} & \ding{51} &          &83.42 & 66.47 & 62.14 & 87.88 & \textbf{60.26} & 60.97 \\
    \ding{51} & \ding{51} & \ding{51} & \ding{51} &\ding{51} &\textbf{83.96} & \textbf{69.39} & \textbf{63.56} & \textbf{88.18} & 60.25 & \textbf{61.02} \\
  \end{tabular}
  \vspace{0.6cm}
  \caption{Components Ablation.}
  
\end{subtable}
\hfill
\begin{subtable}[t]{0.47\textwidth}
  \centering
  \label{fig:scale}
  \includegraphics[width=\linewidth]{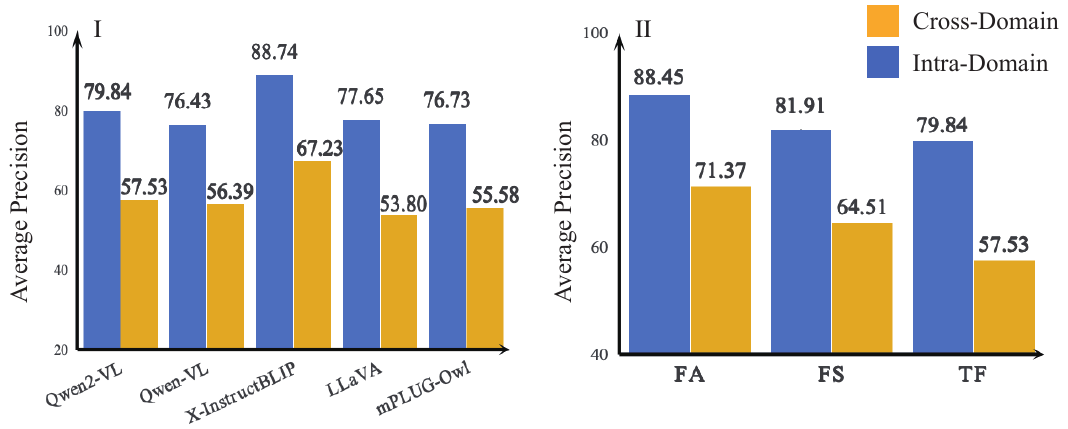}
  \caption{Generalization across MLLMs \& Manip. Perform.}
  
\end{subtable}

\vspace{-0.3cm}

\end{table*}


For the implementation details of all experimental models, the training settings, and the definitions of the evaluation metrics, please refer to the appendix.

\subsection{Quantitative Results}
\noindent\textbf{Effectiveness \& Generalization.} We assess AMD against four SOTA methods on the MDSM and $\text{DGM}^4$ datasets. For MDSM (Tab.~\ref{tab:baselines_Comparison}), we train on \textit{The Guardian} and \textit{NYT}, testing on the rest. For $\text{DGM}^4$ (Tab.~\ref{tab:DGM4_baselines_Comparison}), we train on the largest subset, \textit{The Guardian}. Tab.~\ref{tab:baselines_Comparison} also shows zero-shot results for general-purpose models. Our key findings are:
(1) MLLMs' knowledge boosts performance. Forgery-trace methods like ViLT \citep{ViLT_ICML} and HAMMER series \citep{DGM4_TPAMI} show limited performance, unlike MLLM-based methods like FKA-Owl \citep{liu2024fkaowl} and AMD. For instance, trained on MDSM-NYT (Tab.~\ref{tab:baselines_Comparison}), AMD achieves an 84.47 intra-domain mAP and >60 cross-domain, while HAMMER scores 57.06 and <42, respectively.
(2) AMD achieves strong grounding. AMD attains the best average mIoU of 63.56 (NYT-trained) and 61.02 (Guardian-trained) (Tab.~\ref{tab:baselines_Comparison}). General-purpose models perform poorly (mIoU < 3). AMD's superiority stems from its question-answer heuristic prompts and MOR module, which omits coordinate outputs when no manipulation is detected, thus reducing unnecessary errors.
(3) AMD generalizes effectively. On $\text{DGM}^4$ (Tab.~\ref{tab:DGM4_baselines_Comparison}), AMD outperforms the HAMMER series on all metrics (74.47 ACC, 52.91 mAP, 80.01 $\text{P}_{tok}$, 51.87 mIoU). It also surpasses FKA-Owl in ACC, mAP, and mIoU, despite a lower $\text{P}_{tok}$.

\noindent\textbf{Generalization Assessment across MLLMs.}
To assess generalization on different MLLMs, we evaluated an NYT-trained AMD on test narratives generated by four MLLMs: Qwen-VL \citep{QwenVL}, X-InstructBLIP \citep{xInstructBLIP_2024}, LLaVA \citep{LLAVA_NeurIPS}, and mPLUG-Owl \citep{ye2024mplugowl3longimagesequenceunderstanding}. Results (Tab.~\ref{tab:ablation_and_discussion}b, chart \uppercase\expandafter{\romannumeral1}) show robust performance, with intra-domain (NYT) and cross-domain APs exceeding 76 and 53, respectively.

\noindent\textbf{Details of Manipulation Type Detection.}
Using AMD trained on the NYT domain as an example, the bar chart \uppercase\expandafter{\romannumeral2} in Tab.~\ref{tab:ablation_and_discussion}b shows that text-modal (TF) manipulations are harder to detect than image-modal ones. FA achieves intra-domain AP of 88.45 and cross-domain AP of 71.37, while TF reaches 79.84 and 57.53, respectively. This highlights the deceptive nature of MLLM-generated narratives.

\subsection{Ablation Studies}
\begin{figure*}[!t]
    \centering
    \includegraphics[width=0.9\linewidth]{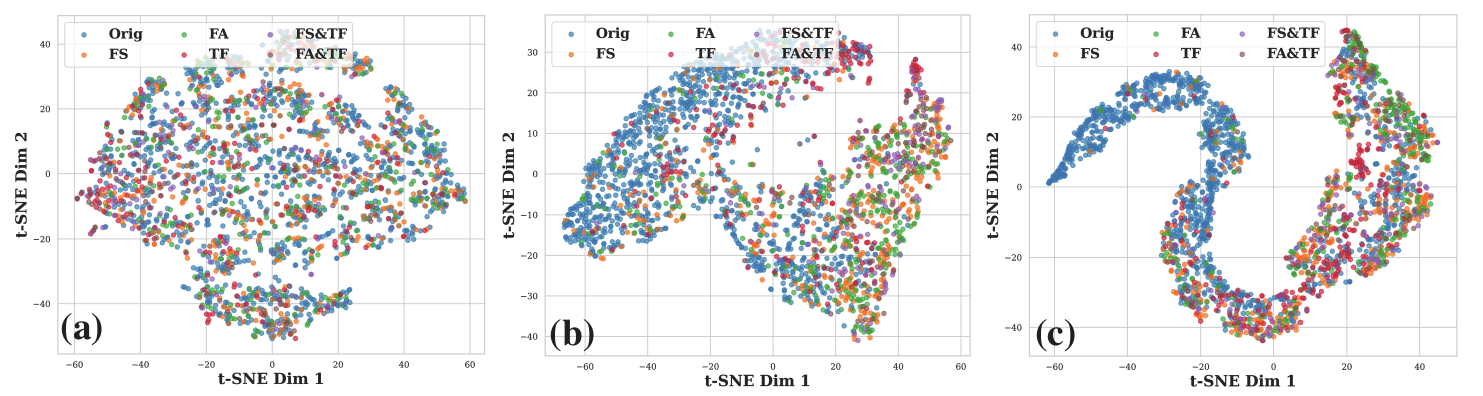}
    \caption{Visualization of Artifact Tokens. (a) Visualization of $E_a$ in $S_{inp}$. (b) Visualization of $\hat{E_a}$ in $S_{a}$. (c) Visualization of $\hat{E_a^m}$ in $S_{m}$.}
    \label{fig:sequence_tsne}
\vspace{-0.5cm}
    
\end{figure*}

\noindent\textbf{Component Ablation.} Tab.~\ref{tab:ablation_and_discussion}a presents the results for each component considered in our study. We use a fine-tuned Florence-2 with our designed prompts as the baseline. As shown, incorporating Artifact Pre-perception Encoding (APE) improves all three task metrics, especially binary classification accuracy, which increases from 76.92 to 82.93 on NYT and from 82.33 to 87.68 on Guardian. This demonstrates that pre-perception of manipulation traces is vital for aiding MLLMs in multi-media manipulation detection. Adding auxiliary tasks, such as Dual-Branch Manipulation (DBM) and Image Manipulation Grounding (IMG), enhances fake news classification and grounding performance, while also slightly improving binary classification. Notably, DBM significantly boosts AMD's mAP, increasing from 47.18 to 66.47 on NYT and from 56.43 to 60.26 on Guardian. Furthermore, the incorporation of the Token Redundancy Penalty (TRP) yields comprehensive performance gains, especially exhibiting stable improvements in ACC and mIoU across both domains. 

\noindent\textbf{Artifact Token Visualization.} Fig.~\ref{fig:sequence_tsne} visualizes the Artifact token at different stages via t-SNE \citep{tsne_JMLR}. As shown in Fig.~\ref{fig:sequence_tsne}a to c, the sample points progressively form more distinct clusters, clearly demonstrating the effectiveness of our AMD optimization in enhancing the Artifact Token's ability to distinguish between different categories.

\noindent\textbf{Efficiency Discussion.} Tab.~\ref{tab:Efficiency_Comparison} compares params scale and throughput (images-text pairs processed per second) on RTX 4090. With 276M parameters, AMD is smaller than FKA-Owl (6,771M), enabling faster training and inference. Among comparable-sized models like ViLT and HAMMER, AMD achieves slower speed than HAMMER but significantly outperforms them on misinformation detection tasks. Overall, AMD delivers strong detection performance while maintaining a compact architecture and efficient inference.

\begin{table}[t]
  \centering
  \caption{Efficiency comparison of multi-modal learning methods.}
  \vspace{-0.1cm}
  \label{tab:Efficiency_Comparison}
  \scriptsize
  \begin{tabular}{@{}lcccc@{}}
    \toprule
    \multirow{2}{*}{Method} 
    & \multicolumn{2}{c}{Params (M) $\downarrow$} 
    & \multicolumn{2}{c}{Throughput (p/s) $\uparrow$} \\[-0.08cm]
    \cmidrule(lr){2-3} \cmidrule(lr){4-5} \\[-0.35cm]
    & Total & Trainable & Train & Inference \\[-0.08cm]
    \midrule
    ViLT & 121.07 & 121.07 & 1.85 & 2.38 \\
    HAMMER(++)\tnote{1} & 441.12 & 228.25 & 28.97 & 61.28 \\
    FKA-Owl & 6771.98 & 33.55 & 1.25 & 1.33 \\
    \rowcolor{gray!20}
    \textbf{AMD} 
      & \textbf{276.95} 
      & \textbf{276.95} 
      & \textbf{5.55} 
      & \textbf{13.38} \\
    \bottomrule
  \end{tabular}
  \vspace{-0.65cm}
\end{table}



\section{Conclusion}
\label{sec:Conclusion}
This study discloses two critical limitations in current multimedia manipulation detection: underestimation of dynamic semantic deception risks posed by MLLMs and the unrealistic, semantically incoherent misalignment artifacts among existing benchmarks. To address these challenges, we construct the MLLM-Driven Synthetic Multimodal (MDSM) dataset and the Artifact-aware Manipulation Diagnosis (AMD) framework to address this new and challenging problem. MDSM contains over 441k semantically aligned image-text pairs from five major media domains, providing a solid foundation for training robust misinformation detection models. AMD integrates Artifact Pre-perception Encoding and Manipulation-Oriented Reasoning to enhance detection of MLLM-generated multimodal disinformation. Comprehensive Experiments demonstrate the framework’s strong generalization and effectiveness against MLLM-driven deception.



\section*{Acknowledgment}
This work was supported by the National Natural Science Foundation of China (NSFC) under Grant No. 62302140, and the National Key Research and Development Program of China under Grant No. 2023YFC3321600. The authors also gratefully acknowledge the support from the Guangdong Basic and Applied Basic Research Foundation (2025A1515012281), the Nanjing Municipal Science and Technology Bureau (202401035), and the University of Macau (MYRG-GRG2024-00077-FST-UMDF).

{
    \small
    \bibliographystyle{ieeenat_fullname}
    \bibliography{main}
}
\clearpage

\makeatletter
\renewcommand{\numberline}[1]{}
\makeatother

\twocolumn[
{
\renewcommand\twocolumn[1][]{#1}  
\maketitlesupplementary

}]


\section{Related Work}
\label{sec:RelatedWork}

\noindent\textbf{Deepfake Detection.}
The rapid progress of generative models and the surge in synthetic content have accelerated advances in Deepfake detection. Existing work spans unimodal and multimodal approaches. Unimodal, including image-based~\citep{ICCV_9710718, CVPR_9157215} and text-based~\citep{ACL_NEP_text_unimodal, huang2023faking} approaches, already achieve strong results. With the rise of Multimodal Large Language Models (MLLMs), multimodal Deepfake detection has received increasing attention~\citep{liu2024mmfakebench, DGM4, liu2024fkaowl}.
Regarding datasets, Shao \textit{et al.}~\citep{DGM4} introduced the pioneering DGM$^4$ benchmark for multimodal manipulation detection and grounding. However, its manipulations are rule-based, leading to semantically fragmented image-text discrepancies that do not accurately reflect real-world misinformation. MMFakeBench~\citep{liu2024mmfakebench} recognized this limitation and proposed generating semantically aligned news images using text-to-image models. Yet such semantically matched samples constitute only 30\% of its fake subset, and the dataset contains merely 11k samples, limiting its utility for training robust detectors. Existing Deepfake datasets also fail to consider the risk of semantically coherent but misleading text generated by modern MLLMs.
On the methodological side, HAMMER~\citep{DGM4} integrates contrastive learning to build a detector capable of classifying manipulation types and grounding manipulated regions, but it does not address cross-domain robustness. Beyond conventional multimodal detectors, FKA-Owl~\citep{liu2024fkaowl} employs a 7B-scale MLLM with several architectural modifications to enhance generalization. However, it is trained on DGM$^4$, where text manipulations follow fixed editing rules rather than being synthesized by MLLMs, making it unsuitable for detecting more subtle, semantically aligned misinformation produced by modern models. Moreover, FKA-Owl performs only binary real/fake classification without fine-grained manipulation type prediction or grounding, and its large backbone and heavy architectural design result in substantially increased computational cost and slower inference.

\noindent\textbf{Multi-Modal Large Language Model.}
In recent years, Multi-Modal Large Language Models have emerged as a crucial technology for understanding and reasoning across multiple modalities, particularly text and images. By extending the capabilities of Large Language Models (LLMs) to incorporate visual inputs, these models have demonstrated outstanding performance in tasks such as image captioning and visual question answering. CLIP ~\citep{CLIP_Radford2021LearningTV} and ALIGN ~\citep{ALIGN_Jia2021ScalingUV} leveraged contrastive learning to align visual and textual representations, enabling efficient zero-shot vision-language understanding. Subsequently, models such as Flamingo ~\citep{Flamingo_NEURIPS2022} and BLIP-2 ~\citep{BLIP2_ICML} have introduced vision-language transformers, integrating pre-trained LLMs with vision encoders to enhance cross-modal reasoning and generative capabilities. More recently, GPT-4V ~\citep{GPT4TR} and Florence-2 ~\citep{Florence2_CVPR} have significantly enhanced the potential of MLLMs in tackling complex multi-modal tasks by leveraging a more efficient framework and larger-scale pre-training data.
A key advantage of MLLMs is their acquisition of extensive world knowledge through large-scale pretraining, which substantially strengthens their reasoning abilities in downstream tasks. Such knowledge not only enhances cross-modal understanding but also proves essential for handling challenging problems, including misinformation detection.



\section{Prompt Paradigm}
\subsection{Prompt for AMD}
The details of the heuristic question-answer prompts in AMD are as follows: 

\noindent\textit{\#\#\#Human:}

\texttt{The following are multiple choice questions about fake news detection. The text caption of the news is: <Text>. The identity and emotion of the face, and the semantic and sentiment of the text should not be manipulated. Question: Is there any face swap/attribute or text fabrication in the news?}

\texttt{A. No.}

\texttt{B. Only face swap.}

\texttt{C. Only face attribute.}

\texttt{D. Only text swap.}

\texttt{E. Both face swap and text fabrication.}

\texttt{F. Both face attribute and text fabrication.}

\texttt{If there is manipulation of a face, locate the one most likely manipulated face in the image and append the results to your selected option.}

\texttt{The answer is:}

\noindent \textit{\#\#\#Assistant:}

\texttt{<Option>[Manipulated face: <loc\_x\textsubscript{1}><loc\_y\textsubscript{1}><loc\_x\textsubscript{2}><loc\_y\textsubscript{2}>]}

\begin{figure*}[t]
    \centering
    \includegraphics[width=0.95\linewidth]{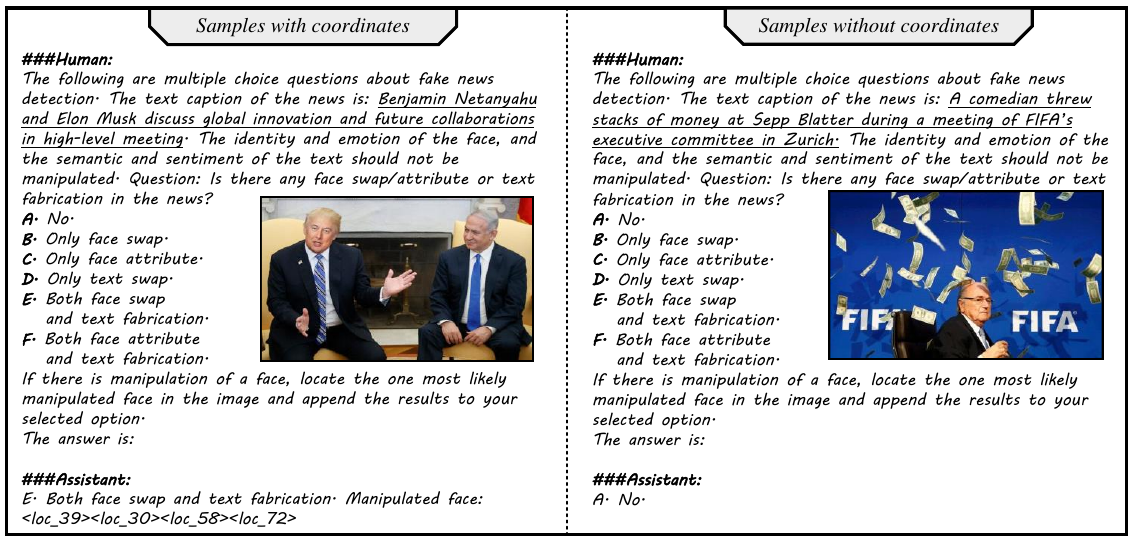}
    \caption{Examples of Image-Prompt pairs in AMD.}
    \label{fig:prompt_demo}
\end{figure*}

Where $<\text{Text}>$ refers to the textual narratives paired with the input image,  $<\text{Option}>$  represents the correct answer option for this sample, such as \textit{E. Both face swap and text fabrication.} And  $<\text{loc}\_>$ is added to the vocabulary as a special token representing coordinates. Fig.~\ref{fig:prompt_demo} shows two kinds of prompts.

\subsection{Prompt for General-purpose Model}

To ensure fairer testing and more credible results for general-purpose models (Tab.2 in the main paper), we enhanced the invocation of general-purpose models by adding more detailed descriptions to the AMD prompt, as follows:

\noindent\textit{\#\#\#Human:}

\texttt{<Same as AMD>}

\texttt{If face manipulation, use rectangular box coordinates in the format of [x1,y1,x2,y2], where the top-left vertex of the image is defined as (0,0) and the bottom-right vertex as (1,1) for relative positioning, and append the results to the option you have selected.}

\texttt{DO NOT output analysis. ONLY output final answer in format: [Option + Coordinates (if applicable).]}

\section{Experimental Setup}
\label{sec:Experimental Setup}
\subsection{Implementation Details}
All experiments are conducted on 4 NVIDIA GeForce RTX 4090 GPUs using \textbf{Distributed Data Parallel (DDP)} training in PyTorch. The image encoder \(\mathcal{E}_v\) is based on DaViT~\citep{ding2022davit}, with Florence-2-B~\citep{Florence2_CVPR} serving as the backbone. The APE \(\mathcal{E}_m^p\) is based on the Florence-2 encoder and remains frozen during training. Thus, in the APE stage, only the artifact token, classifier head, and attention pooling module are jointly trained. The classifiers and bounding box (bbox) detector consist of two Multi-Layer Perceptron layers, with output dimensions of 2 and 4, respectively. For manipulation detection guidance, the \textbf{Dual-Branch Manipulation} shares a common classifier.

The training images are resized to \(224 \times 224\) and undergo random horizontal flipping. The batch size per GPU is set to 6, and the model is trained for 12 epochs. We use the AdamW optimizer~\citep{AdamW_ICLR} with an initial learning rate of \(1e^{-7}\) and a weight decay of 0.01. A cosine learning rate scheduler with a warm-up phase is applied, gradually increasing the learning rate to \(1e^{-6}\) in the first 1000 steps, and then decaying it to \(1e^{-7}\) throughout training. Our code will be released to provide further implementation details.

\subsection{Baselines}
We adapt four state-of-the-art multi-modal methods to the MDSM setting for comparison, including three multi-modal manipulation detection models and one multi-modal learning approach:
\begin{itemize}

\item \textbf{HAMMER} ~\citep{DGM4} is a pioneering model for the multi-modal manipulation detection and grounding. It employs two unimodal encoders to extract visual and textual forgery features, which are then aligned through contrastive learning. Following this, a multi-branch transformer architecture with two specialized decoders is utilized for manipulation detection and grounding. 
\item \textbf{HAMMER++} ~\citep{DGM4_TPAMI} is a more powerful model that builds upon HAMMER by integrating contrastive learning from both global and local perspectives. 
\item \textbf{FKA-Owl} ~\citep{liu2024fkaowl} is another pioneering model designed for large vision-language models to perform multi-modal fake news detection, and it demonstrates outstanding cross-domain performance. Since FKA-Owl does not support fine-grained classification tasks, we fine-tuned it using the same prompts as those used for AMD. 
\item  \textbf{ViLT} ~\citep{ViLT_ICML}, for the multi-modal learning approach, is a representative single-stream method where cross-modal interaction layers operate on the concatenation of image and text inputs. For adaptation, We add classification and detection heads to the corresponding outputs of the model.
\end{itemize}

\subsection{Evaluation Metrics}
To comprehensively evaluate our proposed MDSM, we follow the rigorous evaluation protocols and metrics outlined in~\cite{DGM4} for all manipulation detection and grounding tasks. The detailed evaluation setup is organized as follows:  

\begin{itemize}
  \item \textbf{Binary Classification.} \textbf{Accuracy (ACC)} is adopted as the evaluation metric to measure the correctness of real/fake news classification results.  

  \item\textbf{Multi-Label Classification.} 
  For multi-label classification tasks, we employ the \textbf{mean Average Precision (mAP)}, which measures the per-class average precision and then takes the arithmetic mean across all manipulation types. This macro-averaged mAP provides a comprehensive evaluation of the model’s overall performance across different manipulation types.

  \item\textbf{Manipulated Image Bounding Box Grounding.} To evaluate the precision of predicted manipulated bounding boxes, we calculate the \textbf{mean Intersection over Union (mIoU)} between the ground-truth and predicted coordinates for all testing samples. This metric quantifies the spatial overlap between detected regions and actual manipulated areas, reflecting the localization accuracy of the model.  

  \item\textbf{Manipulated Text Token Grounding.} In the $\text{DGM}^4$ benchmark, an additional task of manipulated text token grounding is included. For this task, \textbf{Precision} is used as the evaluation metric to measure the accuracy of identifying manipulated text tokens within input sequences.  
\end{itemize}

This standardized evaluation framework ensures a systematic and comparative assessment of MDSM across diverse manipulation scenarios, aligning with both general detection tasks and benchmark-specific requirements.

\section{Ethics Statement}

The MDSM dataset and associated analyses were created solely to support research on detection and mitigation of modern MLLM-driven multimodal misinformation. We recognize that assembling realistic, semantically coherent synthetic examples entails dual-use risks: the same materials and procedures could be misused to produce deceptive content. To minimize harm, we adopt a harm-minimizing, controlled-release approach: we will not publish the generation pipeline, detailed prompts, or prompt--response pairs to prevent their exploitation by adversaries for generating harmful content; public distribution is limited to vetted, research-only access under a signed Data Usage Agreement (DUA); distributed images will carry conspicuous visual watermarks and standardized metadata tags; high-fidelity originals and sensitive metadata will be withheld; images of minors and clearly sensitive contemporary conflict content have been excluded; and reserve the right to revoke access on evidence of misuse. Full technical and procedural details of these safeguards are documented in the Appendix and in the dataset README.




\clearpage


\end{document}